# Improving TTS for Shanghainese: Addressing Tone Sandhi via Word Segmentation

Yuanhao Chen
yuanhao.chen.25@dartmouth.edu

## Abstract

Tone is a crucial component of the prosody of Shanghainese, a Wu Chinese variety spoken primarily in urban Shanghai. Tone sandhi, which applies to all multi-syllabic words in Shanghainese, then, is key to natural-sounding speech. Unfortunately, recent work on Shanghainese TTS (text-to-speech) such as Apple's VoiceOver has shown poor performance with tone sandhi, especially LD (left-dominant sandhi). Here I show that word segmentation during text preprocessing can improve the quality of tone sandhi production in TTS models. Syllables within the same word are annotated with a special symbol, which serves as a proxy for prosodic information of the domain of LD. Contrary to the common practice of using prosodic annotation mainly for static pauses, this paper demonstrates that prosodic annotation can also be applied to dynamic tonal phenomena. I anticipate this project to be a starting point for bringing formal linguistic accounts of Shanghainese into computational projects. Too long have we been using the Mandarin models to approximate Shanghainese, but it is a different language with its own linguistic features, and its digitisation and revitalisation should be treated as such.

## 1 Introduction

Shanghainese is a variety of Wu Chinese spoken primarily in urban Shanghai and globally by the Shanghainese diaspora.

Despite its formerly prominent status as a lingua franca in the Yangtze River Delta region, Shanghainese is now a minority language in Shanghai, with Putonghua (Standard Mandarin) being the dominant language in the city. Furthermore, the situation is only exacerbated by the general sentiment among the younger generation that Shanghainese is a "low-status" language, and by their adoption of the linguistic model that one nation should use only one language (Gilliland, 2006). Education plays a crucial role in this process, as this sentiment was mainly observed among college students. As a result, most young people in Shanghai, whether native or an immigrant, are unable to speak Shanghainese fluently (Weng, 2023).

With Putonghua being the perceived authentic and superior language in many aspects of life, crucially including education, it is direly important to preserve the linguistic variety in Shanghai by promoting the use of Shanghainese. Digitisation of a substratum is an effective way to promote the language in teaching, learning, and various other dimensions of cultural life (Villa, 2002).

In this project, I aim to build a TTS (text-to-speech) system for Shanghainese, which is a crucial component in the digitisation of a language, serving as a bridge between digitised written and spoken forms of the language. This is not to say that there is no existing work on Shanghainese TTS. Notably, Apple Inc. (2017) added Shanghainese to the list of languages supported by VoiceOver, the screen reader built into Apple's operating systems. However, the quality of the synthesised speech is not satisfactory, and definitely not on par with the quality of the synthesised speech for other Sinitic languages such as Putonghua. The main problem with Shanghainese VoiceOver is its occasional poor performance with tone sandhi, especially LD (left-dominant sandhi), a suprasegmental phonological process involving a specific bounding domain (Roberts, 2020). For example, the word /[zã²³.he³³⁴]$_{\text{LD domain}}$/ 'Shanghai' has to be pronounced with LD as [zã².he⁴] (the left syllable's rising contour is spread over to the right one).

This paper will explore the possibility of improving tone sandhi in Shanghainese TTS by putting focus on annotating the bounding domain of LD during preprocessing of input texts. Specifically, I will segment the input text into lexical words. My results confirm that this approach is effective in improving the quality of synthesised speech in terms of tone sandhi.

## 2 Methodology

### 2.1 Overview

The key to improving LD in Shanghainese TTS is to annotate the bounding domain of LD. Instead of training a model for this task, which is difficult due to lack of resources, I will perform word segmentation, because lexical words highly correlate with the domains for LD (Kuang and Tian, 2019); formally, LD domains can be formed by the left edges of lexical words, with a few exceptions (Roberts, 2020). Thus, this prosodic annotation can be transformed into word segmentation, giving us the overall pipeline of this paper as shown in Fig. 1.

### 2.2 Datasets

The data basis of TTS models is a list of corresponding audio files and transcriptions. I am using a dataset of an ASR project (Cosmos-Break, 2023), which contains 2,012 audio files and corresponding transcriptions in Chinese characters, totalling 5,607 seconds of speech of a single Shanghainese speaker. The types of speech in the dataset range from single words to phrases and sentences. The audio is resampled to 16 kHz for training.

For word segmentation and phonemisation, we are going to need a phonemically annotated lexicon of Shanghainese. I am using one containing more than 125,000 lexical entries, 51,000 of which have corresponding romanisations (Chen, 2022).

### 2.3 Word Segmentation

The `jieba` library is the most popular open-source Chinese word segmentation library, which comes with a Mandarin dictionary out of the box (Sun, 2023). It implements two models, namely a trie of a deterministic finite automaton (DFA) pre-built from the dictionary, and a hidden Markov model (HMM) with Viterbi algorithm as a backup for unknown words. We will have to rely on the Mandarin dictionary because word-segmentation algorithms require word frequency information, which is not available in the Shanghainese lexicon, but we can patch the model by adding weights to Shanghainese-specific words on top of Mandarin weights. This works well, as written Mandarin and Shanghainese are very similar.

### 2.4 Phonemisation

The aforementioned dictionary (Chen, 2022) is used to romanise segmented words in *Yahwe* Wu Chinese Romanisation (吳語協會式拼音). As the dictionary only contains traditional forms, before romanising, the words are converted to Traditional Chinese using Kuo (2023). Then, `Qieyun` (Mikazuki, 2022) is used to add any necessary tone numbers to the romanisation. The romanisation is then converted to broad IPA transcription largely following the paradigm of Qian (2007). Because the goal is not to accurately transcribe phonetically but to effectively feed the TTS model with phonemic contrasts, some notational techniques are employed to reduce the number of ambiguous digraphs, such as using ⟨c⟩ for /tɕ/ and ⟨j⟩ for /dʑ/.

### 2.5 Training the TTS Model

The TTS model presented in this paper is trained using the VITS end-to-end TTS model (Kim et al., 2021). Compared to previous popular TTS models such as Glow-TTS (Kim et al., 2020), VITS employs a variational autoencoder (VAE) to produce a latent model of the input text and a stochastic duration predictor, which allows the model to express the natural one-to-many relationship in which a text input can be spoken in multiple ways with different pitches and rhythms.

This is perfect for the task that this paper aims to accomplish: The model can effectively learn that the same string in the input may be pronounced with a different pitch depending on whether tone sandhi applies, i.e., depending on the prosodic environment.

The pre-processed text and audio are fed into the VITS model, which is trained for 50K steps with a batch size of 32.

## 3 Experiments

I conducted an $n \times 3 \times 5 \times 4$ experiment, where $n = 11$ is the number of native Shanghainese participants, 3 is the number of speakers generating the audio samples, 5 is the number of sentences, and 4 is the number of metrics. A subjective human evaluation (MOS, mean opinion score) is conducted under each condition on a 1–5 scale, with 5 being the best (see details in Fig. 8).

The three speakers are
(1) The model presented by this paper.
(2) Shanghainese VoiceOver (Apple Inc., 2017).
(3) This paper's author, a native speaker.

The four metrics follow what is proposed by Cardoso et al. (2015):
(1) Comprehensibility: How well can you understand the meaning of the audio?

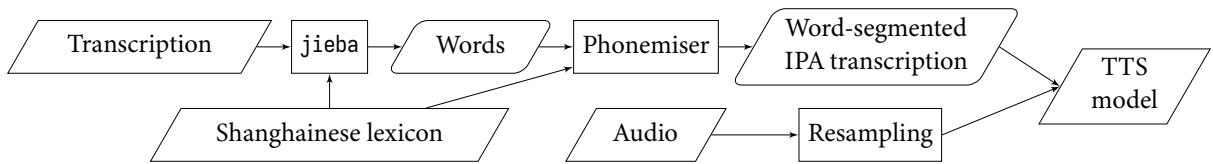

Figure 1: Overview of the pipeline in training the TTS model, with emphasis on the text preprocessing steps.

(2) Naturalness: How natural does the audio sound?
(3) Accuracy: How well does the audio match how a native speaker like you would pronounce it?
(4) Intelligibility: How much effort does it take to make sense of the audio?

## 4 Results

### 4.1 Tone Sandhi production

In all test sentences (containing 23 different LD domains to trigger sandhi), all speakers produce the correct tone sandhi, except for VoiceOver, which fails to produce the correct tone sandhi in sentence 5, 儂弗要弗二弗三個 'don't be nasty'.

弗二弗三個 'nasty' /[vəʔ¹².ɲi²³.vəʔ¹².se⁵¹.ɦəʔ]$_{LD}$/ is a word forming a pentasyllabic LD domain, which should surface tonally as [vəʔ¹.ɲi³.vəʔ².se².ɦəʔ¹], but VoiceOver incorrectly treats it as /[vəʔ¹².ɲi²³]$_{LD}$ [[vəʔ¹².se⁵¹.ɦəʔ]$_{LD}$ …]$_{RD}$/, splitting the word into two LD domains, and assuming that an extra RD (right-dominant sandhi) domain dominates the second LD domain, which is doubly wrong. This produces the incorrect surface form [vəʔ¹.ɲi³ vəʔ¹.se¹.ɦəʔ³]. Compare the pitch in Figs. 6 and 7.

### 4.2 Questionnaire

One out of 11 questionnaires collected is discarded for being incomplete. The valid 10×3×5×4 opinion scores are used for various statistical analyses.

The overall MOS of three speakers are shown in Table 1. Speakers 1 and 2 are not significantly different from each other ($p$ = 0.64); they both have significantly lower scores than speaker 3 ($p \ll 0.001$). A breakdown of scores grouped by metrics shows similar statistical relationships between speakers (see Table 2).

In a breakdown of scores grouped by sentences, two statistically significant differences are found: Speaker 1 has a significantly lower score than speaker 2 in sentence 2, and speaker 2 has a significantly lower score than speaker 1 in sentence 5 (both $p \ll 0.001$; see Table 3).

## 5 Discussion

From the comparison of overall MOS and the comparison of MOS grouped by metrics, we can see that the human acceptability of this model is generally comparable to that of VoiceOver. Statistically significant differences emerge when we break down the scores by sentences, showing that both models have their own strengths and weaknesses.

### 5.1 Shortcomings of this model

As shown in Table 3, this model has a significantly lower score than VoiceOver in sentence 2, 上海是一座國際化大都市 'Shanghai is an international metropolis'.

There are tonal differences between what this model (Fig. 2) produces for 國際化大都市 'international metropolis' /koʔ⁵.tɕi⁵³.ho³³⁴ da²³.tu⁵³.zz̩²³/ and what VoiceOver (Fig. 3) and I (Fig. 4) produce. However, this is not the reason for the low score, because this phrase indeed has two alternative prosodic segmentation possibilities: /[[koʔ⁵.tɕi⁵³.ho³³⁴]$_{LD}$ [da²³.tu⁵³.zz̩²³]$_{LD}$]$_{RD}$/ (what this model goes for) and /[[koʔ⁵.tɕi⁵³.ho³³⁴]$_{LD}$]$_{RD}$ [[da²³.tu⁵³.zz̩²³]$_{LD}$]$_{RD}$/ (what the other speakers go for), depending on whether 'international' and 'metropolis' are grouped together in an RD domain.

The main issue with the pronunciation of this model is that the syllable /koʔ/ is produced too long compared to syllables without /-ʔ/ coda. In Shanghainese, the glottal stop coda is often realised as the shortening of the nucleus (giving [kŏ]), instead of actually pronounced. However, this model produces a syllable clearly longer than others, with a duration of 0.28 s, or 10% of total speech duration, whereas the native speaker produces this with a duration of 0.16 s, or 6% of total speech duration.

This behaviour is possibly due to VITS's makeshift treatment of blanks in speech — VITS inserts a BLANK token between every input character, which supposedly enhances performance in general (Kim et al., 2021). 25% of the duration of /koʔ/ is actually nearly silence at the end of the syllable, a weird pause to have within a word.

**5.2 Advantages of this model**

The model presented by this paper is significantly better than VoiceOver in sentence 5, because it correctly handles LD of /[vəʔ¹².ɲi²³.vəʔ¹².se⁵¹.ɦəʔ]_LD/, which VoiceOver fails (compare Figs. 5 and 6 to the native speaker version in Fig. 7).

This difference is manifested by the mostly correct word segmentation of this model, which outputs vəʔ-ɲi=vəʔ=se1 gəʔ, where hyphens connect syllables in a known word and double hyphens connect those in an inferred word; gəʔ (/ɦəʔ/ when cliticised) is left out but can be inferred from context by the TTS model because it is a common suffix that clings to the previous word. Admittedly, this result might be improved even further if we train a dedicated prosodic annotation model that can detect clitics, rather than just word-segmentation, but the current result is already satisfactory.

The output of VoiceOver is also correct under a low-high pitch accent analysis of Shanghainese (Roberts, 2020), which predicts a pitch contour of LHLLL for this word. However, to sound more tonally natural, one has to correctly formulate LD domains to get the exact pitch contour, which further highlights the importance of accurate prosodic structure in Shanghainese.

**5.3 Ethical considerations of this project**

The nature of TTS models is to mimic human speech as perfectly as possible. With progress in more natural-sounding TTS models such as the one in this paper, it is possible to use TTS models to impersonate other people's voices, which can be used for malicious purposes such as fraud. More concerningly, voice conversion is easy to do with TTS models with VAE such as VITS, which drastically widens the potential scope of malicious use to any person. While driving the digitisation and revitalisation of a minority language like Shanghainese, we should simultaneously be aware of the potential harm that technology can bring, and definitely refrain from any malicious application.

## 6 Conclusion

In this work, I have presented a TTS model for Shanghainese with the novel approach of emphasising bounding domains of tone sandhi, specifically LD, during text preprocessing. Due to lack of material to train a dedicated annotation model, word segmentation is employed as a proxy for this phonological information, which is shown to be effective in improving the tone sandhi quality of the output speech compared to Apple Inc. (2017). Further improvement in performance of timing and pausing may be achieved by switching to a TTS model that handles blanks in speech better.

Beyond just prosody, the significance of this project should be to raise awareness of the importance of a formal linguistic account in every aspect of the development of computational systems regarding Shanghainese. For example, the dataset used in this project is originally for an ASR project (Cosmos-Break, 2023), but the transcription is scattered with 假借 (phonetic loan characters), where a character is used for its Mandarin pronunciation to approximate the "dialectic" pronunciation, likely because the transcriber reads fluently only in Mandarin. For example, 萨 (a surname, Mandarin /sa/) is used for 啥 ('what', Shanghainese /sa/); such practice is common but greatly hinders a consistent and formal treatment of Shanghainese orthography and lexicon in computational systems, as the character used to approximate varies from person to person, and the phenomenon itself is a manifestation of Mandarin centralism which marginalises Shanghainese.

Specific to the topic of this project, the lack of a computational model implemented as per a formal linguistic account of Shanghainese tone system is a major obstacle to the improvement of tonal performance in TTS. Even word segmentation, which is a makeshift solution of prosodic annotation, is carried out by the makeshift approach of using the Mandarin–pre-trained `jieba` model.

In general, despite the low-resource status of most substrata, it is important to be alerted that over-reliance on resources of the superstratum is only a makeshift solution that can both undermine the authenticity of the result and take the focus away from the development of the scaffolding (implementation of formal linguistic accounts) of substrata. Therefore, to engineer reliable and minority-friendly computational systems, further research should really put the development of the scaffolding of substrata at the top of the agenda.

## References

Apple Inc. 2017. VoiceOver. Apple Inc.

Walcir Cardoso, George Smith, and Cesar Garcia Fuentes. 2015. Evaluating text-to-speech synthesizers. In *Critical CALL – Proceedings of the 2015 EUROCALL Conference, Padova, Italy*, pages 108–113. Research-publishing.net.

Yuanhao Chen. 2022. Rime Yahwe Zaonhe. Zenodo.

Cosmos-Break. 2023. Shanghainese ASR.

Joshua Gilliland. 2006. *Language Attitudes and Ideologies in Shanghai, China*. Ph.D. thesis, The Ohio State University.

Jaehyeon Kim, Sungwon Kim, Jungil Kong, and Sungroh Yoon. 2020. Glow-TTS: A Generative Flow for Text-to-Speech via Monotonic Alignment Search.

Jaehyeon Kim, Jungil Kong, and Juhee Son. 2021. Conditional Variational Autoencoder with Adversarial Learning for End-to-End Text-to-Speech.

Jianjing Kuang and Jiapeng Tian. 2019. Tone Representation and Tone Processing in Shanghainese. In *Proceedings of the 19th International Congress of Phonetic Sciences*, Melbourne, Australia.

Carbo Kuo. 2023. OpenCC (Open Chinese Convert 開放中文轉換).

Ayaka Mikazuki. 2022. Qieyun-python. nk2028.

Nairong Qian. 2007. *Shanghai Fangyan (Shanghainese)*, first edition. Haipai Wenhua Congshu. Wenhui Press, Shanghai.

Brice David Roberts. 2020. *An Autosegmental-Metrical Model of Shanghainese Tone and Intonation*. Ph.D. thesis, UCLA.

Junyi Sun. 2023. Jieba.

Daniel J. Villa. 2002. Integrating technology into minority language preservation and teaching efforts: An inside job. *Language Learning & Technology*, 6(2).

Shihong Weng. 2023. The second generation of "New Shanghainese": Their language and identity. *San Diego Linguistic Papers*, 12.

## A  Tables

| Speaker | Overall MOS |
|---|---|
| 1 | 4.14 ± 0.12 |
| 2 | 4.19 ± 0.14 |
| 3 | 4.83 ± 0.06 |

Table 1: Overall MOS of three speakers. Confidence interval: 95%; same for tables below.

| Speaker | Accuracy | Comprehensibility | Intelligibility | Naturalness |
|---|---|---|---|---|
| 1 | 4.06 ± 0.26 | 4.48 ± 0.17 | 4.36 ± 0.22 | 3.66 ± 0.28 |
| 2 | 4.02 ± 0.36 | 4.58 ± 0.18 | 4.38 ± 0.24 | 3.76 ± 0.32 |
| 3 | 4.82 ± 0.11 | 4.82 ± 0.11 | 4.86 ± 0.10 | 4.82 ± 0.14 |

Table 2: MOS of three speakers by metrics.

| Speaker | Sentence 1 | Sentence 2 | Sentence 3 | Sentence 4 | Sentence 5 |
|---|---|---|---|---|---|
| 1 | 4.35 ± 0.22 | 3.70 ± 0.34 | 3.85 ± 0.28 | 4.28 ± 0.25 | 4.53 ± 0.24 |
| 2 | 4.45 ± 0.24 | 4.70 ± 0.18 | 4.15 ± 0.26 | 4.55 ± 0.24 | 3.08 ± 0.41 |
| 3 | 4.83 ± 0.12 | 4.82 ± 0.12 | 4.88 ± 0.13 | 4.75 ± 0.16 | 4.88 ± 0.11 |

Table 3: MOS of three speakers by sentences.

## B  Figures

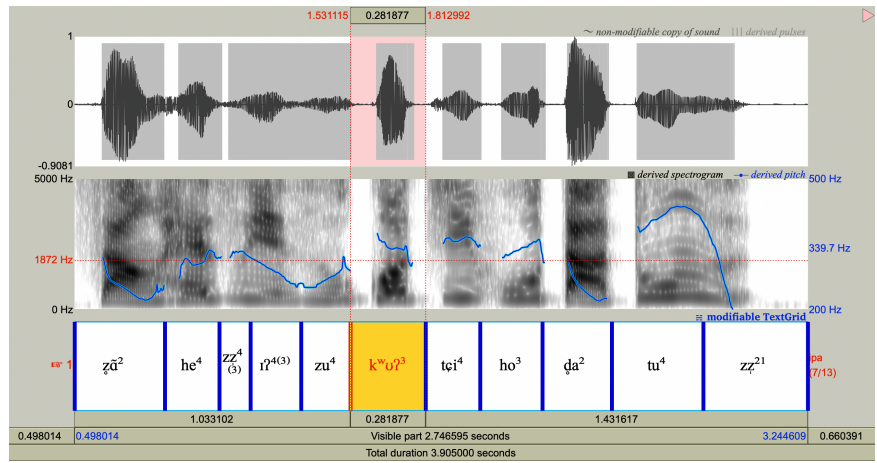

Figure 2: Sentence 2 by speaker 1 with broad phonetic annotation. Tones are represented by Chao tone letters as their phonetic realisations. Parentheses indicate uncertain tone height that is within acceptable range and not crucial to the analysis. Same for figures below.

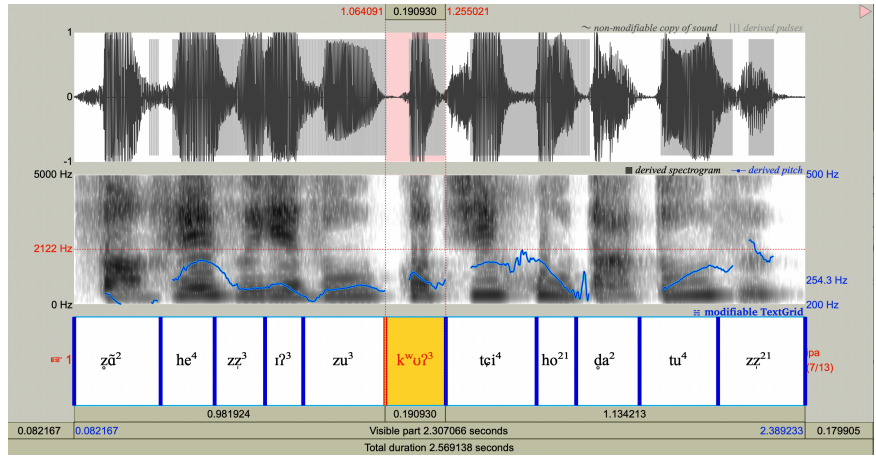

Figure 3: Sentence 2 by speaker 2.

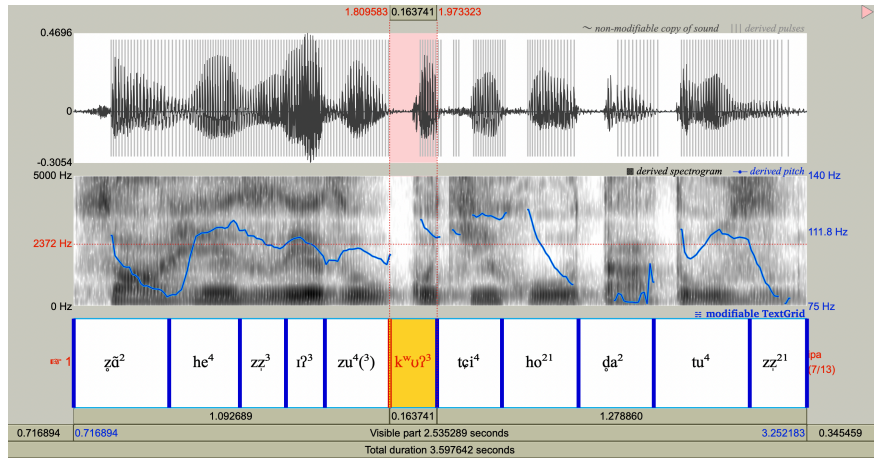

Figure 4: Sentence 2 by speaker 3.

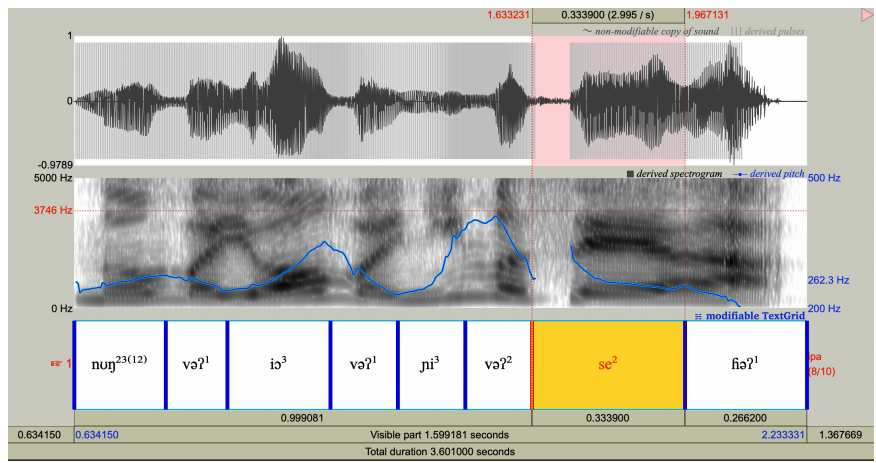

Figure 5: Sentence 5 by speaker 1.

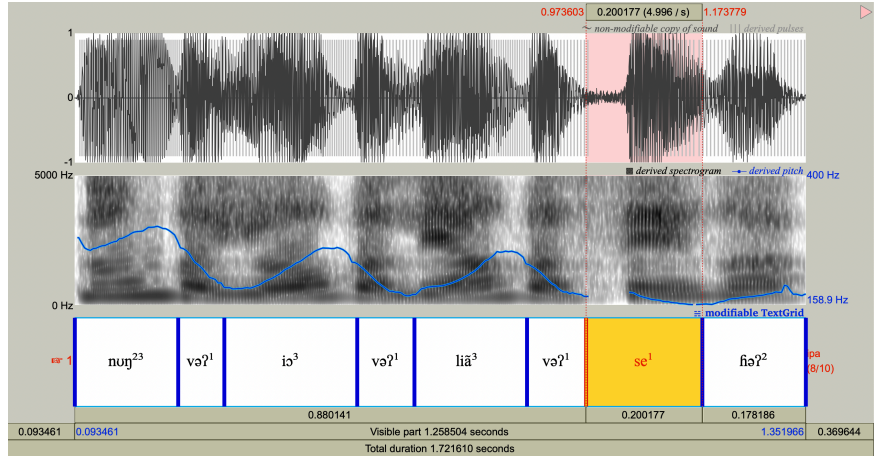

Figure 6: Sentence 5 by speaker 2.

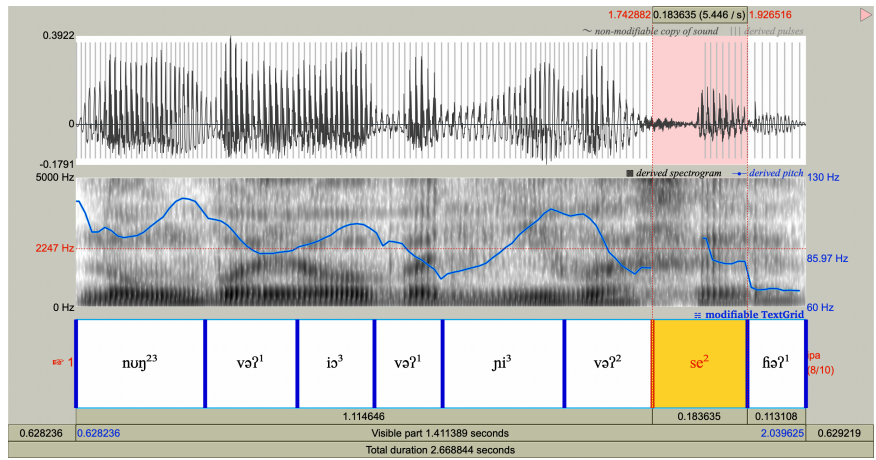

Figure 7: Sentence 5 by speaker 3.

## C  Questionnaire

![Questionnaire]

Figure 8: Questionnaire.